\documentclass{article}

\usepackage{PRIMEarxiv}
\usepackage[utf8]{inputenc} 
\usepackage[T1]{fontenc}    
\usepackage{hyperref}       
\usepackage{url}            
\usepackage{booktabs}       
\usepackage{amsfonts}       
\usepackage{nicefrac}       
\usepackage{microtype}      
\usepackage{lipsum}
\usepackage{fancyhdr}       
\usepackage{graphicx}       
\graphicspath{{media/}}     
\usepackage{tikz}
\usepackage{enumitem}

\title{Co-TAP: Three-Layer Agent Interaction Protocol Technical Report

}

\author{
  Co-TAP Team \\
  \url{https://github.com/ZTE-AICloud/Co-TAP} \\
  \texttt{Email:AIM@zte.com.cn} \\
}

\begin{document}
    
    
        
        
        
        
\noindent\includegraphics[width=3.5cm]{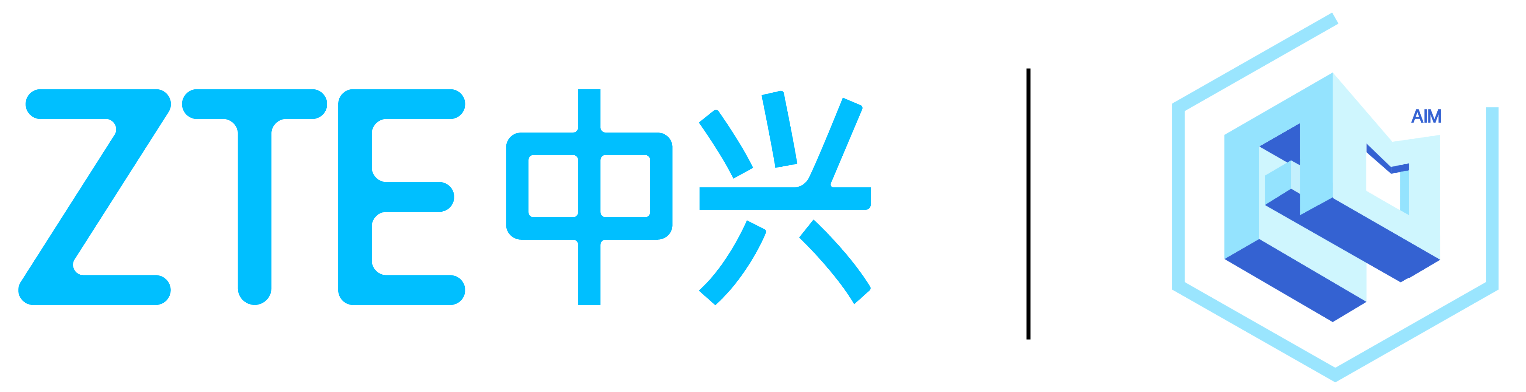}
\vspace*{-0.2cm}
\maketitle
\begin{abstract}
This paper proposes Co-TAP (T: Triple, A: Agent, P: Protocol), a three-layer agent interaction protocol designed to address the challenges faced by multi-agent systems across the three core dimensions of Interoperability, Interaction and Collaboration, and Knowledge Sharing. Current multi-agent systems widely suffer from the ``information silo'' phenomenon. The lack of unified communication protocols among heterogeneous agents leads to high adaptation costs. Interactions between humans and agents, as well as between agents themselves, lack standardized mechanisms, resulting in the redundant transmission of contextual information and inefficient collaboration. More importantly, the experience and knowledge of agents are difficult to accumulate, transfer, and share, which restricts the emergence of collective intelligence. To address the aforementioned challenges, we have designed and proposed a layered solution composed of three core protocols: the Human-Agent Interaction Protocol (HAI), the Unified Agent Protocol (UAP), and the Memory-Extraction-Knowledge Protocol (MEK). HAI focuses on the interaction layer, standardizing the flow of information between users, interfaces, and agents by defining a standardized, event-driven communication paradigm. This ensures the real-time performance, reliability, and synergy of interactions. As the core of the infrastructure layer, UAP is designed to break down communication barriers among heterogeneous agents through unified service discovery and protocol conversion mechanisms, thereby enabling seamless interconnection and interoperability of the underlying network. MEK, in turn, operates at the cognitive layer. By establishing a standardized ``Memory (M) - Extraction (E) - Knowledge (K)'' cognitive chain, it empowers agents with the ability to learn from individual experiences and form shareable knowledge, thereby laying the foundation for the realization of true collective intelligence. We believe this protocol framework will provide a solid engineering foundation and theoretical guidance for building the next generation of efficient, scalable, and intelligent multi-agent applications.
\end{abstract}


\section{Introduction}

\subsection{Research Background and Motivation}
The proliferation of Large Language Models (LLMs) has driven a paradigm shift in the field of artificial intelligence, giving rise to advanced autonomous agents capable of complex reasoning, planning, and tool use~\cite{xi2025rise,wang2024survey,hao2023reasoning}. These agents are increasingly being deployed in Multi-Agent Systems (MAS) to solve complex, distributed problems that exceed the capabilities of a single agent~\cite{krishnan2025advancing,mushtaq2025harnessing}. The core value of MAS lies in decomposing cognitive loads through a ``division of labor and collaborative debate'' approach, achieving ``collective intelligence'' that surpasses individual capabilities through ``human-like collaboration'' and ``emergent behaviors''. This model is not merely an aggregation of computational power but a fundamentally new problem-solving paradigm. From collaborative software development~\cite{wasif2025multi,deloach2005multiagent}, scientific discovery~\cite{zhang2025collective,ghareeb2025robin}, to complex business process automation~\cite{tebourbi2025bpmn,sulis2022multi} and social simulation~\cite{li2008agent,davidsson2000multi}, MAS are poised to become the cornerstone of future intelligent applications. Meanwhile, the explosive growth in demand for ``plug-and-play, open, and interconnected collaboration'' of agents in communication scenarios such as AIOps and 6G autonomous networks further highlights the central role of MAS in the intelligent transformation of communication infrastructure. However, the evolution from single-agent systems to large-scale, heterogeneous multi-agent systems has exposed significant engineering and theoretical challenges, hindering their large-scale deployment~\cite{mascardi2019engineering,yan2025beyond}. Existing MAS architectures often suffer from redundant data transmission when handling interactions among users, user interfaces, and agents, leading to unnecessary network bandwidth and computational overhead. This makes it difficult to efficiently execute session states across different agents, exacerbating the problems of interoperability and high adaptation costs. As the diversity and complexity of agents continue to increase, the lack of a standardized interaction and collaboration framework leads to ecosystem fragmentation (i.e., the ``information silo'' phenomenon)~\cite{lazaridou2020emergent,ouyang2025code2mcp}. This fragmentation manifests as several key bottlenecks that are the focus of this paper.

\subsection{Engineering Bottlenecks in Multi-Agent Systems}
Despite their promising prospects, building robust and scalable multi-agent systems still faces three major engineering obstacles:
\begin{itemize}
\item \textbf{Redundant Context Transmission in Interactions:} In current MAS architectures, the lack of standardized communication mechanisms for interactions among users, user interfaces, agents leads to inefficient and redundant transmission of contextual information~\cite{qasim2024effective}. For example, session state, user identity, and task history may be repeatedly bundled with every message, consuming extra bandwidth and increasing processing overhead~\cite{cemri2025multi,julian2004developing}. Furthermore, the absence of a standardized event-driven framework makes it difficult to build truly collaborative environments that enable seamless, reliable, and real-time interactions between multiple agents and human users~\cite{menda2018deep,meyer2014event}.
\item \textbf{Lack of Interoperability and High Adaptation Costs:} The modern agent ecosystem is highly heterogeneous. Agents are developed using different frameworks (e.g., LangChain~\cite{calvaresi2017challenge}, AutoGen~\cite{wasif2025multi}, CrewAI~\cite{venkadesh2024unlocking}), run on diverse platforms, and communicate via various protocols (e.g., Agent-to-Agent Protocol (A2A)~\cite{jeong2025study}, Message-oriented Middleware Communication Protocol (MCP)~\cite{cao2025large}, or proprietary RESTful APIs~\cite{gorodetsky2020system}). The absence of a unified communication standard forces developers to create costly and fragile point-to-point integrations for each new agent or service~\cite{ehtesham2025survey,hosseini2025role}. The resulting N-to-N adaptation problem not only inhibits scalability but also increases system fragility—a change in any single agent's protocol can trigger global cascading effects~\cite{baldoni2020fragility,jang2004efficient}. The recent emergence of multiple protocols further underscores the urgent need for a unified, open standard, a problem that has yet to be effectively addressed.
\item \textbf{Difficulties in Knowledge and Experience Sharing:} While individual agents can learn from their own interactions and ``memory''~\cite{wang2023cooperative,dorri2018multi}, a core challenge in MAS is the inability to effectively share these learned experiences across the collective~\cite{ghafarollahi2025sciagents}. Agent memories are often stored in unstructured or semi-structured formats (e.g., conversation logs, vector databases), containing vast amounts of experiential data~\cite{banares2005multi}. However, without a standardized process for abstracting valuable insights (knowledge) from raw memory, experiences remain isolated within individual agents~\cite{yu2012group,su2001logical}. This prevents the emergence of true collective intelligence, forcing each agent to learn from scratch and repeat similar mistakes, thereby limiting the overall system's performance and adaptability~\cite{liu2022distributed,baldoni2023accountability}. Fundamentally, the current problem is that the rapid advancement of LLM agents in areas like reasoning and tool use has outpaced the development of their underlying communication and memory infrastructure.
\end{itemize}

\subsection{Core Protocol Framework and Objectives}
We propose a three-layer protocol framework designed to provide a comprehensive solution for building next-generation multi-agent systems. This framework comprises three innovative protocols: the Human-Agent Interaction (HAI) Protocol, the Unified Agent Protocol (UAP), and the Memory-Extraction-Knowledge (MEK) Protocol.

\begin{itemize}
\item \textbf{HAI Protocol (Human-Agent Interaction Protocol):} HAI is an open, lightweight, and event-based standardized protocol that acts as a translator and bridge between backend agents and frontend UIs. It aims to standardize the interaction design, message structure, event handling, and transmission mechanisms between UIs and agents to ensure efficient, standardized, and bidirectional real-time communication. Built on an event-driven architecture, HAI uses a unified JSON event stream format. Its core features include support for token-by-token real-time streaming, full snapshot or incremental synchronization of agent states (state sharing), support for agents to invoke frontend tools (tool orchestration), and enabling users to communicate with and control agents throughout their entire lifecycle (e.g., start, pause, stop). The ultimate goal is to enhance the fluency and user experience of human-agent interactions.
\item \textbf{UAP Protocol (Unified Agent Protocol):} The UAP Protocol adopts a core strategy of ``modular decomposition and ecosystem-based construction'' to address critical bottlenecks in MAS deployment, such as the inability of agents to perform autonomous discovery, difficulties in implementing complex tasks, high cross-protocol adaptation costs due to the coexistence of multiple protocols, and extra overhead from redundant context transmission. By establishing a unified multi-protocol service registry and gateway, UAP decomposes multi-agent collaboration capabilities into independent functional modules, such as an AI Gateway, registration and discovery, identity authentication, and communication protocols. Its primary objective is to create an ``Internet of Agents'', breaking down silos between agents, providing a unified service plane and governance capabilities, supporting mainstream agent protocols (e.g., A2A, ACP, MCP), and lowering application integration barriers through protocol standardization and unified adaptation. This accelerates application deployment while seeking an effective balance between system reliability, scalability, and security.
\item \textbf{MEK Protocol (Memory-Extraction-Knowledge Protocol):} The MEK protocol is a standardized memory management framework designed for agents, defining a logical chain: M (Memory Core) → E (Value Extraction) → K (Knowledge Distillation). This involves three key aspects: first, Memory, which addresses how to efficiently and structurally store and organize multi-modal memory information to build long-term experience repositories; second, Extraction, which focuses on how to intelligently and accurately extract high-value information snippets from vast memories based on task requirements; and third, Knowledge, which concerns how to distill and generalize extracted, personalized information into reusable and shareable general knowledge across agents. Its core value lies in establishing a standardized process for memory processing and knowledge sharing, enabling agents to transform isolated experiences into systematic cognitive abilities, thereby achieving true intelligence emergence.
\end{itemize}

The three protocols proposed in this paper are closely related to existing research directions but also offer significant extensions. Previous studies have conducted in-depth explorations in areas such as human-computer collaboration and trust modeling~\cite{julian2004developing,menda2018deep,meyer2014event,wang2023cooperative,dorri2018multi,yousefli2020maintenance}, agent communication languages and standardization~\cite{davidsson2000multi, mascardi2019engineering, yan2025beyond,ouyang2025code2mcp,sarkar2025survey}, middleware interoperability and IoT/IoRT frameworks~\cite{zhang2025collective, ghareeb2025robin, tebourbi2025bpmn, sulis2022multi, li2008agent,newton2021scalability}, agent memory and knowledge sharing~\cite{jeong2025study, cao2025large, gorodetsky2020system, ehtesham2025survey}, and communication mechanisms in multi-agent reinforcement learning~\cite{lazaridou2020emergent,calvaresi2017challenge, partalas2008hybrid, ding2024learning, lee1998stability}. However, these works often focus on a single layer, such as communication semantics, interaction patterns, or knowledge storage, and lack a unified protocol system that spans infrastructure, interaction, and cognition. Therefore, this research is the first to propose and systematize the HAI, UAP and MEK protocols, aiming to lay a scalable, trustworthy, and intelligent protocol foundation for the large-scale deployment of multi-agent systems.

\subsection{Report Structure}
The structure of this report is organized as follows:

Chapter 1: This chapter first introduces the research background and motivation of this technical report, clarifying the significance of the research. It then describes the three major engineering bottlenecks faced by multi-agent systems, which leads to the introduction of the three core technical protocols proposed in this report.

Chapter 2: HAI Human-Agent Interaction Protocol: This chapter provides a detailed description of its role as a bridge between the frontend and agents, its unified event stream architecture, core event types (e.g., lifecycle, business data, and tool call events), and the transmission mechanisms (e.g., SSE) that support real-time, collaborative interactions.

Chapter 3: UAP Unified Agent Protocol: This chapter focuses on its core strategy of ``modular decomposition and ecosystem-based construction'', its layered registration architecture, the gateway's protocol conversion mechanism (e.g., MCP to A2A/ACP), and the multi-modal registration and high-availability design under a unified service model.

Chapter 4: MEK Memory-Extraction-Knowledge Protocol: This chapter elaborates on the M-E-K logical chain, the four core memory mechanisms, the four-step process of the Extraction protocol, and how to achieve multi-agent knowledge sharing and absorption through a standardized ``KnowledgeItem''.

Chapter 5: Protocol Collaboration and Synergy: This chapter analyzes how HAI, UAP, and MEK work together synergistically in complex task scenarios.

Chapter 6: Conclusion and Future Outlook: This chapter summarizes the core value and technical benefits of the proposed protocol architecture and provides an outlook on the future development directions of agent technology.

\begin{figure}[h]
\centering
\includegraphics[width=0.4\linewidth]{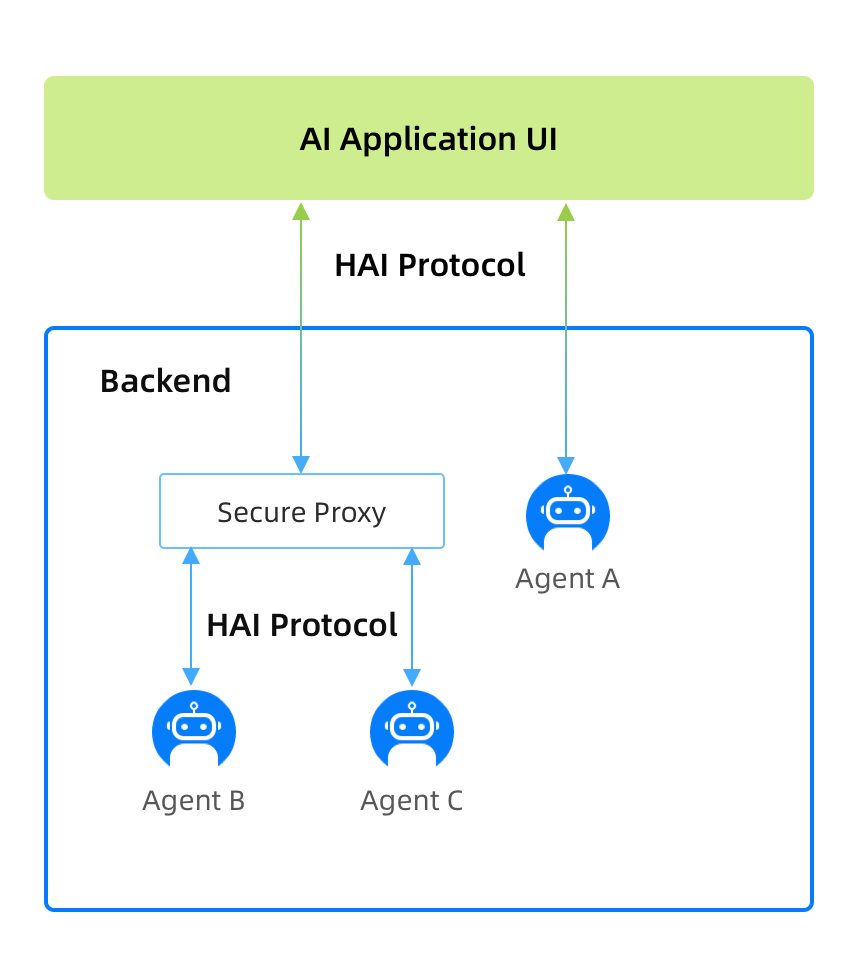}
\caption{HAI Architecture Diagram}
\label{fig1}
\end{figure}

\section{HAI: Human-Agent Interaction Protocol}

\subsection{Design Philosophy and Core Objectives}

The HAI (Human-Agent Interaction) protocol is an open, lightweight, and event-driven standardized communication protocol designed to establish an efficient, reliable, and scalable interaction channel between backend intelligence (Agents) and frontend user interfaces (UI). It not only serves as the core bridge connecting AI capabilities with the user perception layer but also plays a key role in the system by unifying semantic expression, ensuring real-time responsiveness, and supporting dynamic interactions. The HAI architecture is illustrated in Figure \ref{fig1}.

\subsubsection{Unified Interaction Language}

The core design philosophy of HAI is to establish a universal ``interaction language''. This language is based on a structured JSON event stream, where all interactive behaviors—from task initiation to tool calls, and from state synchronization to error feedback—are expressed through predefined event types. By introducing a unified event stream mechanism, HAI achieves ``decoupling'' between the frontend and backend: the frontend does not need to understand the internal logic of the Agent and only needs to render or respond based on the event type and data structure; similarly, the backend does not need to concern itself with the specific implementation details of the frontend and only needs to output the event stream according to the specification. This standardization significantly reduces the cost of adapting to multiple frameworks and enhances the system's portability and maintainability.

\subsubsection{Real-time and Streaming Experience}

Modern users have increasingly high expectations for interaction response speed, especially in generative AI scenarios, where users expect to see content ``rendered as it is generated'' rather than waiting for the complete result to be returned. To this end, the HAI protocol natively supports a token-by-token streaming push mechanism, ensuring that the content output by the Agent can be delivered to the frontend in real-time with the finest granularity. This capability relies on the support of underlying transport protocols (such as SSE). Combined with the various streaming events defined by HAI, the frontend can implement highly immersive interactive experiences like character-by-character rendering, typewriter effects, and dynamic progress bar updates. More importantly, streaming not only enhances the user experience but also provides the technical foundation for advanced features such as interruption control and early intervention. Users can pause, modify, or terminate a task at any time during the Agent's generation process, achieving true ``Human-in-the-Loop'' (HITL) collaboration.

\subsubsection{Collaboration and Full Lifecycle Control}

HAI focuses not only on the one-way transmission of information but also emphasizes two-way collaboration between humans and agents. Its design goal is to empower users with initiative and control throughout the entire lifecycle of an Agent task, covering key stages such as initiation, execution, monitoring, intervention, and termination. Specifically, HAI supports collaborative capabilities such as initiation control, state query, pause and resume, forced stop, process tracing, and feedback. This full lifecycle control capability allows HAI to transcend the traditional ``question-and-answer'' interaction model and move towards a deeper, more trustworthy interaction paradigm. The user is not just a recipient of information but also a collaborator and supervisor in the task execution process.

\subsubsection{State and Dynamic UI Generation}

In complex agent application scenarios, static UIs can no longer meet the demands of dynamic task flows. To address this, HAI introduces a state-sharing mechanism and the concept of Generative UI (Gen UI), enabling the frontend to dynamically generate or adjust UI elements based on the Agent's output.

State synchronization can be achieved in two ways: (1) Upon connection establishment or reconnection, the Agent sends a complete state snapshot to the frontend, containing all contextual information of the current session to ensure state consistency between the frontend and backend. (2) During runtime, only incremental state changes are pushed to reduce network overhead and improve efficiency. Based on this state information, the frontend can build real-time updated visualization components such as state panels, task progress graphs, and tool call chain diagrams. Furthermore, HAI supports a Gen UI mode, where UI elements are directly driven and generated by the Agent's output.

\subsection{Key Technical Components}

The technical implementation of the HAI protocol relies on multiple collaborating functional events. Each module has clear responsibilities and well-defined interfaces, collectively supporting the stable operation of the entire interaction system.

\subsubsection{Event Stream Architecture and Events}

HAI is built on an event-driven architecture, abstracting all interactive behaviors into the fundamental communication unit of an ``event''. Each event represents an observable action or state change that occurs during the Agent's execution, and it is encapsulated in a structured JSON format and pushed to the frontend through a streaming channel.

The event stream has the following characteristics:
\begin{itemize}
\item Orderliness: Events are pushed sequentially in the order they occur, ensuring the frontend can accurately reconstruct the execution flow.
\item Parsability: Each event contains a clear type field (`type`), facilitating the routing of processing logic on the frontend.
\item Extensibility: Supports custom event types to meet the needs of specific business scenarios.
\item Traceability: Each event can be associated with various contextual identifiers, supporting cross-event tracking.
\end{itemize}

Based on functionality, HAI defines six core event types:

\textbf{Lifecycle Events} This module is used to mark the overall execution cycle of an Agent task and serves as the core basis for the frontend to determine the task's status. Typical events include the start, completion, or error state of a task.

\textbf{Business Data Events} This module manages the core business content interaction between the Agent and the user, primarily used in basic conversational scenarios. Its event stream follows a three-part ``start-content-end'' structure.

\textbf{Tool Call Events} When an Agent needs to call an external tool, this module provides a complete description of the calling process. The event sequence is as follows: (1) declare the intention to call a specific tool; (2) stream the call parameters; (3) the tool call concludes, possibly with the result or reason for failure. This module is a key technical support for implementing Human-in-the-Loop (HITL).

\textbf{Agent-to-Agent Message Events} In multi-agent collaborative systems, a primary Agent may need to communicate with other auxiliary Agents. This module is used to describe such internal collaboration processes, including the start and end of task division and coordination among multiple Agents.

\textbf{State Events} To ensure state consistency between the frontend and backend, especially in scenarios with unstable networks or page refreshes, HAI provides a dedicated state synchronization mechanism, such as sending the complete current state object, sending only the changed parts of the state, or providing a historical snapshot of all messages in the current conversation.

\textbf{Custom and Passthrough Events} To meet specific business requirements, HAI allows developers to define extended events: (1) User-defined event types, which must follow naming conventions, allowing the frontend to execute specific logic based on the type. (2) Passthrough of raw data, used for debugging or transmitting non-standard format information, which typically does not participate in the regular rendering process.

\subsubsection{Message Structure Specification}

This module defines the basic structure for all HAI messages, ensuring format consistency and compatibility. It classifies messages by different sender roles (e.g., user, assistant, system, tool, developer), serving as the core data carrier for interaction and containing the complete transmitted content.

\subsubsection{Transport Mechanism}

The HAI protocol provides a standardized HttpAgent client that supports multiple transport mechanisms, allowing developers to choose the optimal solution based on their specific scenarios.
\textbf{Server-Sent Events (SSE)} SSE is the recommended primary transport method for HAI, suitable for Agent-led, unidirectional streaming communication scenarios. Its advantages include: (1) Based on HTTP long-polling, it offers good compatibility and avoids the complex handshake of WebSockets. (2) It supports continuous pushing of event streams from the server to the client, naturally fitting the requirements of token-by-token stream generation. (3) The message format is simple, making it easy to parse and debug.

\textbf{HTTP Binary Transport}\quad For high-performance, low-latency, or large file transfer requirements, HAI supports an HTTP-based binary encoding transport mechanism. This method uses efficient serialization formats like Protocol Buffers or MessagePack to compress data, significantly reducing network overhead. It is suitable for: (1) transmitting multimedia data (images, audio, video); (2) high-frequency state synchronization; (3) optimized communication on mobile devices or in weak network environments.
\begin{figure}[h!]
\centering
\includegraphics[width=0.6\linewidth]{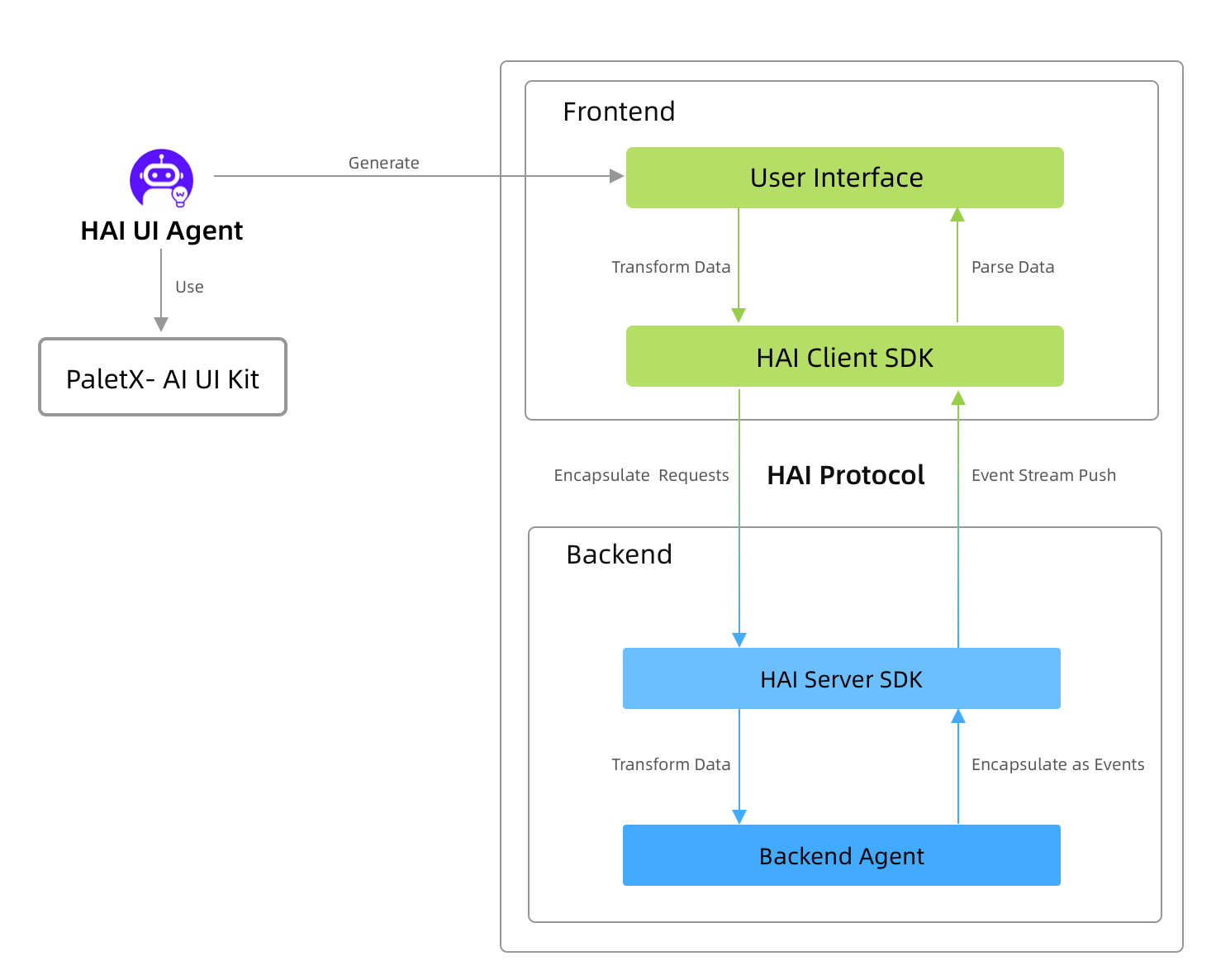}
\caption{HAI Workflow}
\label{fig2}
\end{figure}

\subsubsection{Workflow}

This module defines the standardized, event-driven communication process. The flow begins with the frontend sending a request, the backend then starts the Agent to process it, and subsequently sends a continuous stream of events to the frontend via the SSE protocol. Finally, the frontend receives these events and updates the user interface in real-time. The specific workflow is illustrated in Figure \ref{fig2}.

\section{UAP: Unified Agent Protocol}
\begin{figure}[t]
\centering
\includegraphics[width=0.8\linewidth]{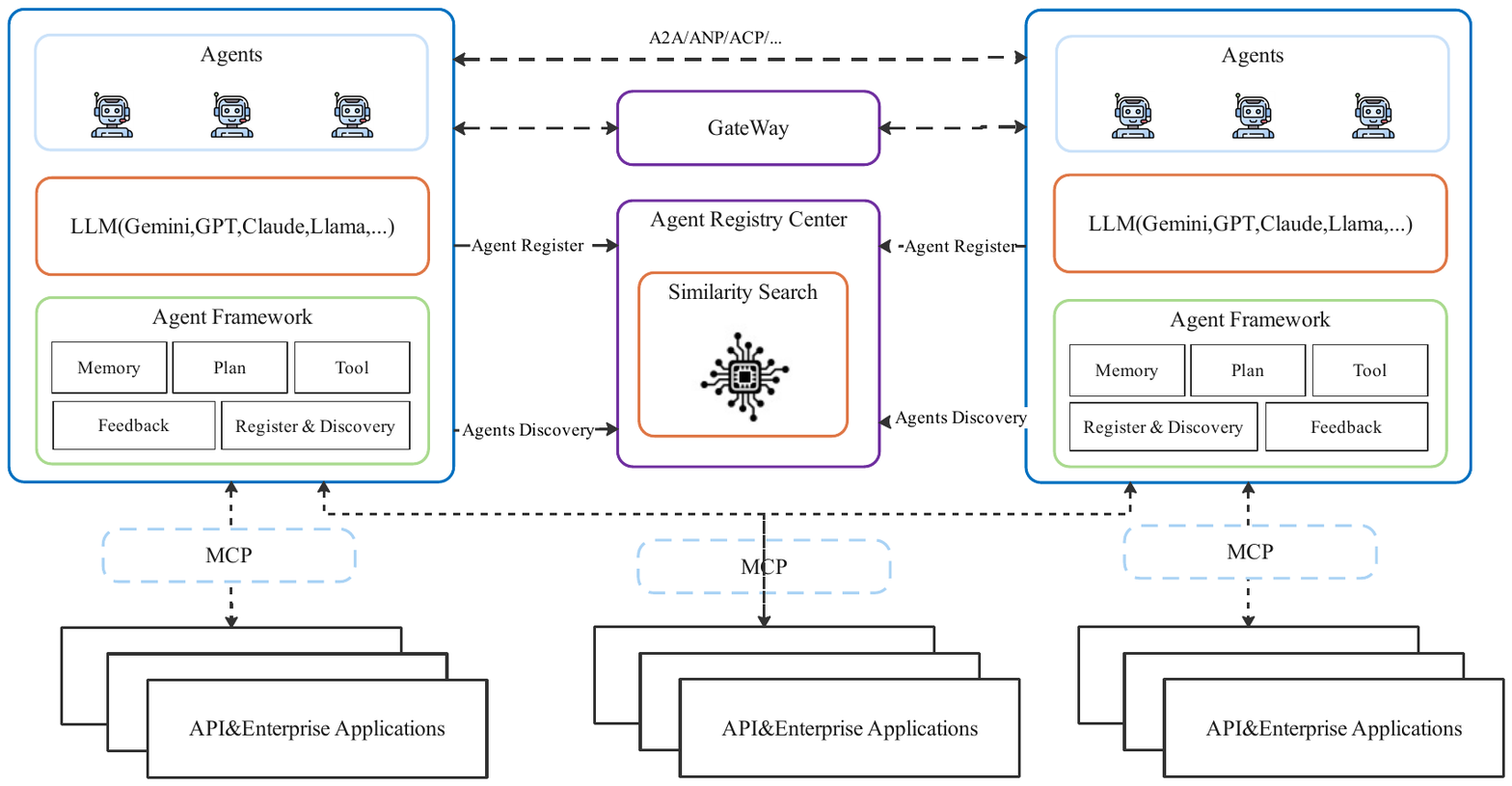}
\caption{UAP Unified Agent Protocol Architecture Diagram}
\label{fig4}
\end{figure}

The Unified Multi-Protocol Service Registry and Gateway (Unified Agent Protocol, UAP) is the cornerstone of this agent technology framework. It is designed to address key bottlenecks in Multi-Agent Systems (MAS), such as the inability of agents to perform autonomous discovery, the high cost of cross-protocol adaptation due to the coexistence of multiple protocols, and redundant context transmission. It also unifies the application lifecycle management process, provides a universal registration mechanism, enhances trust, and promotes secure cross-protocol interactions through a common discovery scheme. The architecture of the Unified Agent Protocol is shown in Figure \ref{fig4}. UAP adopts a core strategy of modular decomposition and ecosystem-based construction, breaking down multi-agent collaboration capabilities into independent functional modules such as an AI Gateway, Registration and Discovery, and Identity Authentication, thereby enhancing the system's flexibility and maintainability.

\subsection{Registration and Discovery}
The UAP Registry Center is a core component of service governance. As the infrastructure for building dynamic collaborative networks, it is responsible for maintaining the full lifecycle status of services and providing a unified registration and discovery mechanism for agents and MCP Servers.

\subsubsection{Design Principles and Unified Service Model}

\begin{figure}[h]
\centering
\includegraphics[width=0.6\linewidth]{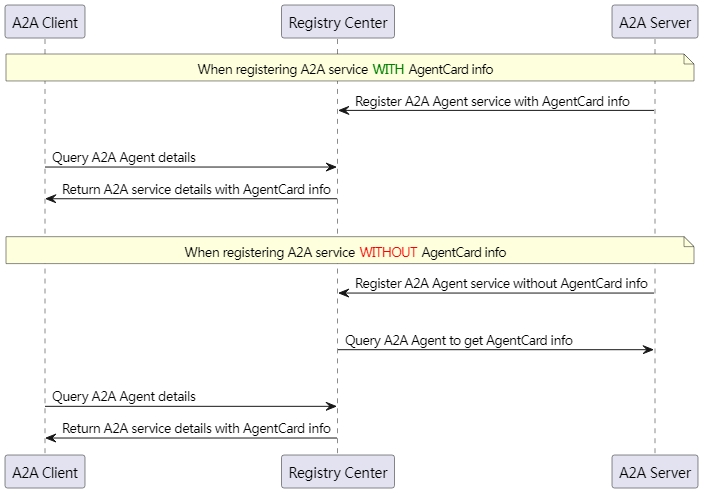}
\caption{A2A Service Architecture Diagram}
\label{fig5}
\end{figure}

The design of the Registry Center follows three core principles to ensure compatibility with heterogeneous agent protocols:
\begin{itemize}
\item Unified Protocol Abstraction: All agent and tool protocols are mapped to a unified service model.
\item Protocol-Specific Extensions: Data unique to each protocol (e.g., the Agent Card for A2A) is mapped to proprietary extension fields to maintain information integrity. Taking the A2A service as an example, the process is shown in Figure \ref{fig5}.
\item Data Consistency: A distributed AP model is adopted to ensure the high availability of the system.
\end{itemize}

The unified service model includes general service information (such as service name, address, protocol type, and version number) as well as protocol-specific extension information (such as Agent Card information for A2A services, Agent Manifest information for ACP services, and MCP Server information for MCP services). Additionally, it supports enhanced search functionalities like fuzzy matching and semantic queries.

\subsubsection{Service Registration and Lifecycle Management}

The UAP Registry Center supports two types of services to accommodate different agent deployment and management needs:

   Ephemeral Services (Self-Registration): Suitable for temporary or on-demand services. These services proactively report their status and require periodic renewal (default renewal interval is 15 seconds, initial TTL is 60 seconds). If a service is not renewed before the timeout, it is automatically deleted. Service information is stored only in memory.

   Persistent Services (Proxy-Registration): Suitable for long-running, resident services, registered by a third-party proxy. The Registry Center proactively and periodically checks the service status. Service information is persisted, and in case of failure, the service is marked as unhealthy but is not automatically deleted.

The Registry Center manages the service lifecycle, including processes such as initial registration, renewal, expiration, deletion, and failure handling. Regarding service capability information, the Registry Center supports services providing this information upon registration (typically for self-registered services), or the Registry can query the service for its capabilities and supplement the information after registration (typically for non-self-registered services).
\subsubsection{Agent Discovery and Routing Management}

Agent discovery is achieved through the Registry Center. Clients (service consumers) can query for the address and capabilities of a target service using criteria such as service name, protocol type, tags, or service description. UAP uses the Registry Center component to manage service routing. The Gateway is configured to use the Registry Center as its data source and forwards requests to services based on the service routes.

Architecturally, UAP employs a hierarchical registration structure. A root registry manages subordinate registry servers, forming a tree-like structure. This allows agents to register their service capability metadata (including functional descriptions, input/output formats, QoS metrics, etc.) through a unified interface. This hierarchical architecture also supports a centralized management mechanism for hierarchical registration, where lower-level registries synchronize their registration information with upper-level registries.
\subsubsection{High Availability Implementation}

The High Availability (HA) of the Registry Center is achieved through three mechanisms:

\begin{itemize}
    \item Cluster Deployment: The Registry Center is deployed with multiple instances to avoid single points of failure.
    \item Data Consistency: Data consistency and high availability are ensured through a backend distributed database (e.g., Consul).
    \item Leader Election: The cluster uses the Raft algorithm to elect a leader. The leader is responsible for data synchronization, coordination, and write operations. Leader election is implemented using a distributed lock based on Consul's session mechanism.
\end{itemize}

\subsection{UAP Gateway Design and Protocol Conversion}

The Gateway is the network traffic entry point and intermediate conversion layer in the UAP framework. It is responsible for protocol conversion, traffic management, route forwarding, and security protection, ensuring seamless integration between different protocols and the overall collaborative capability of the system.
\subsubsection{Design Principles}

The Gateway adheres to the following core principles to ensure the system's flexibility, reliability, and semantic accuracy:
\begin{itemize}
\item Protocol-Agnostic Principle: Ensures the architecture is not tied to any specific protocol standard.
\item Semantic Consistency Principle: The conversion process must preserve the intent of the original request.
\item Minimal Intervention Principle: Protocol conversion is performed only when protocols are mismatched and it is necessary.
\end{itemize}

\subsubsection{Protocol Conversion Mechanism}

The core value of the Gateway lies in solving the communication challenges between heterogeneous agents. It eliminates protocol discrepancies between clients and servers (such as different protocol types, different versions of the same protocol, or different transport protocols) through the following mechanisms:

\begin{enumerate}[label=(\arabic*)]
    \item Bidirectional Adapter Pattern: The Gateway uses the adapter pattern to handle protocol differences. A dedicated bidirectional adapter is created for each pair of protocols (e.g., an A2A-to-MCP adapter).
    \item Adapter Composition: An adapter consists of a semantic mapping table (defining conceptual correspondences) and a conversion rule library (implementing the specific conversion logic). Support for a new protocol only requires adding a new adapter module, without modifying the core system.
    \item Three-Step Process: The implementation of an adapter follows a three-step process: ``Input Parsing -> Semantic Conversion -> Output Generation''. Input parsing converts the original message into an internal representation; semantic conversion applies the mapping rules; and output generation serializes the result into the target protocol format.
    \item Plugin Chain Processing: The Gateway uses a plugin chain mechanism for request handling. The protocol conversion plugin finds and executes the corresponding adapter based on the request's routing protocol and the service's protocol (e.g., finding an ACP-to-A2A adapter).
\end{enumerate}
\begin{figure}[h]
\centering
\includegraphics[width=0.8\linewidth]{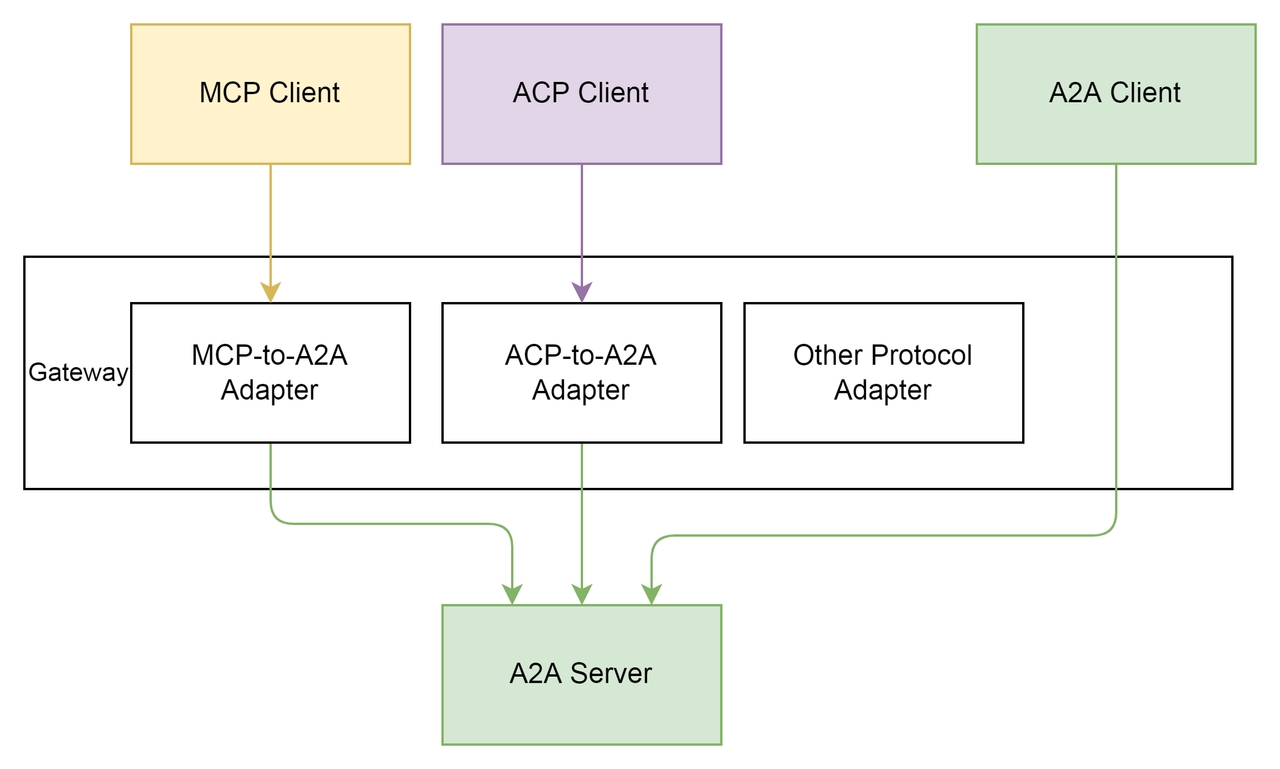}
\caption{Flowchart of an ACP Client Accessing an A2A Service}
\label{fig6}
\end{figure}
\subsubsection{Heterogeneous Protocol Routing and Adaptation}

Using the example of an ACP Client accessing an A2A Server, we can illustrate how the Gateway performs protocol conversion, as detailed in Figure \ref{fig6}. The Gateway supports converting protocols to common interface specifications like HTTP/REST and gRPC, enabling legacy systems to connect to the agent internet through adapters and achieve progressive modernization.

In specific protocol conversion scenarios, for example, the Gateway supports mapping Agent services (like A2A/ACP) to MCP service methods. In this case, the Gateway encapsulates the Agent information as a ``resource'' and the Agent's task execution interface as a `Tool`, thereby mapping MCP interface calls to Agent interfaces. The Gateway also undertakes heterogeneous protocol conversion and can, in conjunction with the hierarchical registration and centralized management mechanism, centrally control protocol conversion plugins to reduce the workload of agent adaptation.

The UAP Gateway plays a critical role in the protocol negotiation phase of the communication flow, deciding whether to use direct communication (when protocols are compatible) or gateway-mediated relay (when protocols are mismatched) to ensure the interoperability of heterogeneous agent services.

\section{MEK: Memory-Extraction-Knowledge Protocol}

The MEK (Memory-Extraction-Knowledge) Protocol is the cognitive and learning layer within the Co-TAP agent protocol architecture. By defining a standardized process, it enables agents to efficiently manage experiences, extract valuable information, and ultimately achieve knowledge sharing and absorption among the collective.

\subsection{Protocol Objectives and the M-E-K Core Logic Chain}

As a core specification for a new generation of agent cognitive architecture, the MEK (Memory-Extraction-Knowledge) Protocol is dedicated to addressing the fundamental bottlenecks in current artificial intelligence systems regarding experience accumulation, knowledge evolution, and collective collaboration. Its design aims to break through the limitations of traditional agents ``one-shot learning and static response'' model, building an intelligent ecosystem with continuous learning capabilities, dynamic adaptability, and cross-agent knowledge synergy.

To realize this vision, the MEK Protocol proposes a clear and executable three-stage core logic chain: M (Memory) $\rightarrow$ E (Extraction) $\rightarrow$ K (Knowledge), forming a complete closed loop from raw perception to advanced cognition. This logic chain not only defines the sequence of information processing but also embodies the cognitive leap of an agent from ``raw experience'' to ``generalizable knowledge''.

\begin{itemize}
\item M (Memory): The Memory module focuses on how to efficiently, structurally, and traceably store and organize multimodal experience information to build an agent's long-term experience repository. This ``memory'' is not merely a data cache but a structured representation of experience processed with semantic understanding and contextual association. It covers various input forms such as text, images, interaction behaviors, and environmental states, encoding them through a unified data model to ensure temporal coherence and spatial retrievability~\cite{hosseini2025role, ghafarollahi2025sciagents, banares2005multi}.
\item E (Extraction): The Extraction stage serves as the critical bridge connecting raw memory with higher-order knowledge. Its core task is to accurately and intelligently identify and extract high-value information fragments from the vast memory pool based on the current task objective, context, or external requests. This process involves more than just keyword matching or vector similarity calculation; it integrates capabilities like semantic understanding, importance assessment, and contextual awareness.
\item K (Knowledge): Knowledge refinement is the culminating stage of the MEK Protocol, aiming to distill and generalize individualized, context-dependent memory content into reusable, shareable, and universal knowledge that can be applied across agents. This process emphasizes depersonalization, standardization, and transferability, enabling the successful experiences of one agent in a specific scenario to be understood, absorbed, and applied by other agents in different but similar situations~\cite{su2001logical, liu2022distributed, baldoni2023accountability, el2009multi}.
\end{itemize}

\begin{figure}[t]
\centering
\includegraphics[width=0.6\linewidth]{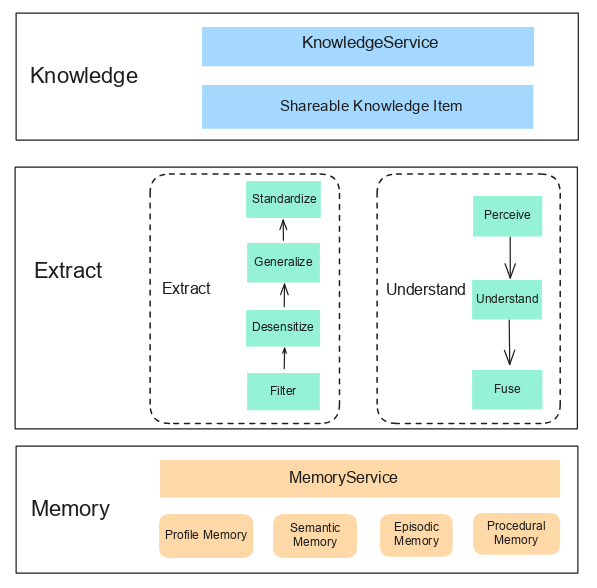}
\caption{MEK Protocol Module}
\label{fig7}
\end{figure}
\subsection{Core Mechanisms and Module Design of Memory}

The Memory module is the cognitive foundation of the MEK Protocol, responsible for the core functions of an agent's experience storage, organization, and management. To cope with complex and variable application scenarios and growing data volumes, the MEK Protocol adopts a classified, layered, structured, and dynamically evolving memory management system. This ensures that memory is both rapidly accessible and durably preserved, while also being highly maintainable and interpretable. The MEK module is illustrated in Figure~\ref{fig7}.

\subsubsection{Memory Storage and Layered Architecture}

The MEK Protocol divides memory into three logical layers, corresponding to different usage frequencies, access speeds, and persistence requirements, forming an efficient memory pyramid structure:

        \textbf{Active Memory Layer} This layer stores high-priority memory items that are highly relevant to the current task or have been frequently accessed recently. These memories are typically in a ``hot state'' and can be invoked by the agent in near real-time, ensuring low-latency responses during interactions. The lifecycle of active memory is short, usually tied to a session or task cycle, and may be demoted to long-term memory or automatically cleared upon task completion. The design of this layer draws inspiration from human working memory, emphasizing immediate availability and context sensitivity.

        \textbf{Long-term Memory Layer} Serving as the ``permanent archive'' of an agent's experience, the long-term memory layer is responsible for the stable preservation of historical interaction records, key events, important facts, and successful patterns with lasting value. Long-term memory supports efficient indexing mechanisms and version control, ensuring the integrity and traceability of information.

        \textbf{Compressed Memory Layer} This layer stores knowledge summaries generated by refining, generalizing, and abstracting raw memories. This content no longer retains specific details but exists in the form of rules, patterns, and strategies. Compressed memory is optimized for rapid reasoning and high-level decision-making, enabling the agent to make efficient judgments without accessing the full memory, thereby significantly improving the system's response speed and generalization capabilities.

Furthermore, the MEK Protocol defines four specialized memory types, each designed to model a specific kind of experience, ensuring accuracy and semantic consistency in information representation:

        \textbf{Workflow Memory} Used to store procedural knowledge for executing specific tasks, such as operational steps, business processes, and automation scripts. This type of memory emphasizes sequence, conditional branches, and execution outcomes, serving as the core basis for an agent to complete complex tasks.

        \textbf{Profile Memory} Used to build cognitive profiles of specific entities (e.g., users, devices, organizations), containing attributes such as preferences, habits, capabilities, and relationship networks. Profile memory supports dynamic updates and multi-perspective fusion, enabling the agent to deliver highly personalized services.

        \textbf{Semantic Memory} Responsible for managing concepts, facts, and their interrelationships, forming the foundation of the agent's internal knowledge graph. Semantic memory is not context-dependent but stores general knowledge, such as ``an apple is a fruit'' or ``HTTP is an application layer protocol''. It supports reasoning, classification, and associative queries, providing crucial support for the agent's understanding of the world and logical inference.

        \textbf{Episodic Memory} Used to record specific events and experiences with rich context, such as the entire process of a user complaint or a failed execution attempt. Episodic memory preserves details like time, place, participants, and causal chains, supporting retrospective analysis and experience summarization. It is a key resource for the agent to learn from failures and identify abnormal patterns.

\subsubsection{Memory Generation Mechanism}

Memory generation follows a three-stage process of Perception $\rightarrow$ Understanding $\rightarrow$ Storage, efficiently converting external inputs into internal memory.

        \textbf{Perception Stage}\quad The agent receives various input signals from the environment through multimodal interfaces, including natural language dialogues, visual images, voice commands, sensor data, and API call results. The perception module is responsible for the initial parsing and formatting of this heterogeneous data, converting it into a unified intermediate representation to lay the groundwork for subsequent understanding.

        \textbf{Understanding Stage}\quad In this stage, the system employs technologies such as Natural Language Processing (NLP), Computer Vision (CV), intent recognition, and entity extraction to deeply analyze the semantics and context of the input. Key tasks include identifying critical entities and events, determining emotional sentiment, inferring user intent, and detecting potential conflicts or anomalies. The results of this understanding directly influence the classification, prioritization, and association-building of the memory.

        \textbf{Storage Stage}\quad The understood information is then structurally encapsulated and categorized into the appropriate memory type based on its content features. The storage process adheres to the following key rules:
        
\begin{enumerate}[label=(\arabic*)]
    \item Deduplication and Merging: Avoid storing duplicate or highly similar information.
    \item Dynamic Prioritization: Automatically adjust the priority of a memory based on the importance of the event or user feedback.
    \item Association Building: Proactively identify and establish logical connections between different memory items to form a memory network.
    \item Conflict Detection and Fusion: When adding new memory, the system should detect potential conflicts with existing memories and prioritize a fusion (Merge) strategy (e.g., merging workflow steps or updating facts) to achieve continuous memory enhancement.
\end{enumerate}

\subsection{Extraction: The Value Extraction Protocol}

Value Extraction (the E stage) is the bridge connecting vast raw memories with shareable knowledge. It defines a rigorous process for distilling personalized experiences into universal knowledge.

\subsubsection{Memory Retrieval Mechanism}

Memory retrieval is the initial filtering stage of the value extraction process. The protocol supports a variety of flexible retrieval types, including exact retrieval for precise lookups based on keywords or unique identifiers, semantic retrieval for matching semantically similar content based on vector similarity, temporal retrieval for tracing back related memory sequences along a timeline, and associative retrieval for performing associative lookups through connections between memory items.

\subsubsection{Standardized Extraction Process}

Value extraction is defined within a core service interface and follows a strict four-step process to refine retrieved raw memories into standardized knowledge units:

\begin{enumerate}[label=(\arabic*)]
    \item Filtering: Identify memory content with universal value from the retrieval results, such as reproducible successful workflows.
    \item Anonymization: Anonymize and clean the filtered memories, stripping all personally identifiable information (PII) or environment-specific sensitive details.
    \item Generalization: Refine the anonymized memory data into a more general knowledge representation that can be applied to a wider range of scenarios.
    \item Standardization: Encapsulate the information into a standard knowledge representation format, making it a unified, independent, and complete knowledge unit ready for cross-agent sharing.
\end{enumerate}

\subsection{Knowledge Sharing and Absorption}

Knowledge sharing is central to the MEK Protocol's goal of achieving collective intelligence. It defines the complete process of how knowledge is transmitted, internalized, and fused among agents.

        \textbf{Structure of Shareable Knowledge} To ensure consistency and interpretability when knowledge is transmitted across agents, all shared knowledge must be encapsulated in a standardized structure. This structure includes a unique identifier for the knowledge, a concise description, structured content, the source memory type, and provenance information such as the identifier of the agent that originally shared it.

        \textbf{Knowledge Sharing and Absorption Process} The sharing agent first executes the complete value extraction process to generate a list of standard knowledge structures and sends it to the target agent. Upon receiving this knowledge structure, the receiving agent understands the knowledge and performs conflict detection against its own memory repository. If a conflict exists, it executes a fusion logic. Finally, the knowledge is converted into a new memory item and stored in its own memory repository, completing the internalization process. During the fusion stage, the system compares the semantic consistency of the new knowledge with existing memories. If a conflict is found (e.g., two different versions of a workflow), a fusion strategy is initiated to generate a superior, integrated solution. Ultimately, the knowledge is transformed into a local ``MemoryItem'' and integrated into the agent's cognitive system.

\section{Protocol Collaboration and Synergy}
The core advantage of the Co-TAP agent protocol architecture lies in the close synergy of its three core protocols—HAI (Interaction and Feedback Layer), UAP (Infrastructure and Interoperability Layer), and MEK (Cognitive and Learning Layer)—in complex task scenarios. These three protocols are functionally complementary, jointly supporting the entire lifecycle of an agent, from task initiation, execution, and interaction to knowledge consolidation.

\subsection{Overall Architecture of Protocol Synergy}
When executing complex tasks, an agent relies on these three protocol layers in sequence, forming a logically clear collaborative loop:
\begin{itemize}
\item HAI (Interaction Layer): Provides real-time visibility and control interfaces for the task execution process, connecting human intelligence with backend collaborative flows to ensure trustworthy and synergistic interaction.
\item UAP (Infrastructure Layer): Provides the necessary interoperability environment for multi-agent collaboration, addressing issues of agent discovery, communication, and protocol adaptation.
\item MEK (Cognitive Layer): Responsible for transforming task execution experiences driven by UAP and HAI into reusable knowledge assets, driving the continuous advancement of collective intelligence.
\end{itemize}

\begin{figure}[t]
\centering
\includegraphics[width=1\linewidth]{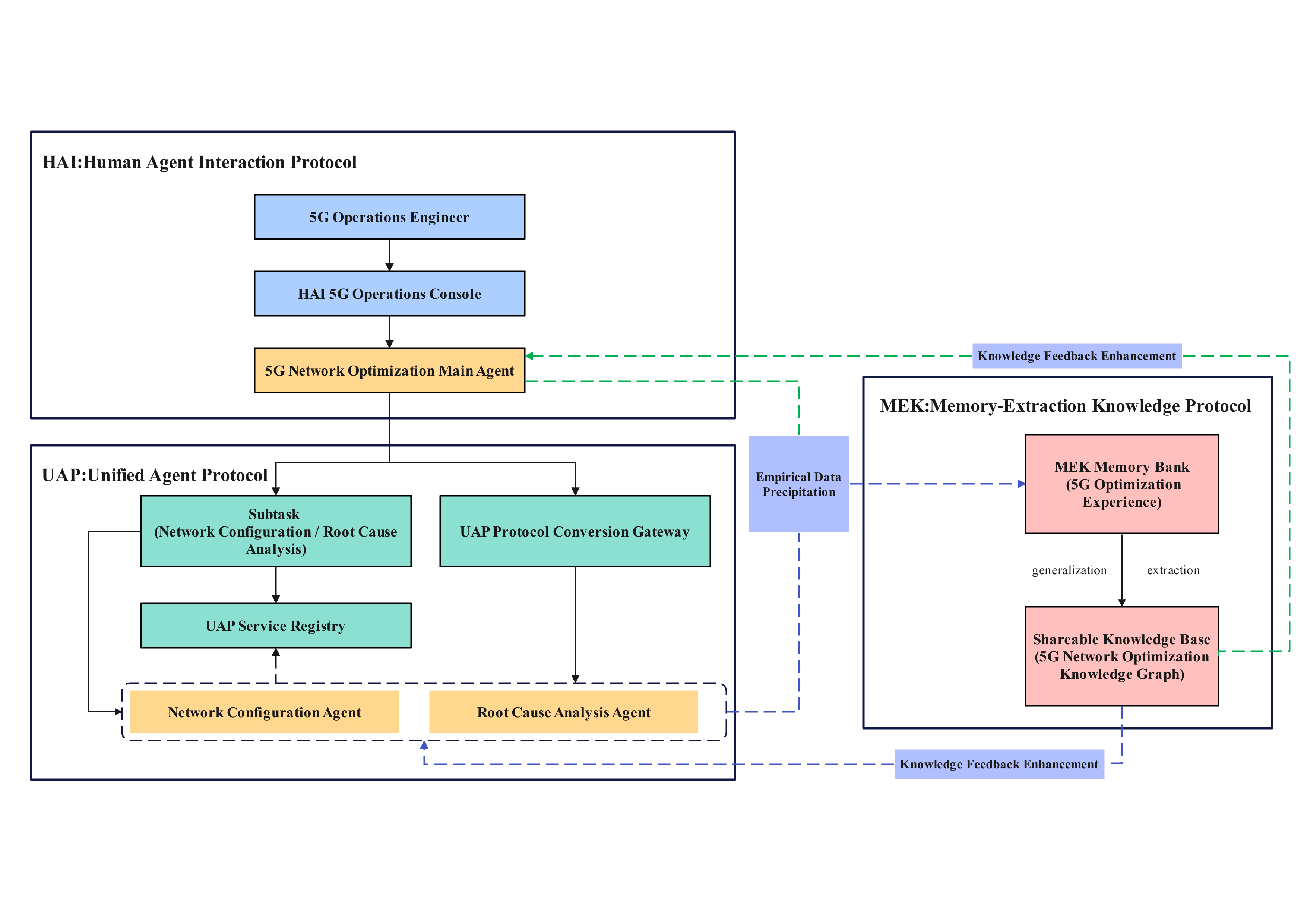}
\vspace{-1.8cm}
\caption{Example of Co-TAP Protocol Collaboration}
\label{fig8}
\end{figure}

\subsection{The Collaborative Loop and Realization of Collective Intelligence}

As shown in Figure \ref{fig8}, we take the operation and optimization scenarios of 5G networks as an example. When a downlink rate anomaly occurs at a base station in a dense urban area, the operation engineer initiates a task through the HAI protocol-based operation console. The request is encapsulated as a business data event and forwarded to the master agent. The master agent then discovers the root cause analysis and network configuration agents via the UAP registry and completes cross-domain or cross-protocol invocation through the UAP gateway. The root cause analysis agent identifies ``PCI conflict-induced interference'' as the root cause, while the network configuration agent generates and deploys a PCI optimization solution. Both agents store their experience into the memory bank using the MEK protocol. The MEK layer extracts this experience, processes it through ``screening - desensitization - generalization - standardization'', and generates knowledge to be stored in the shared knowledge base, thereby enhancing the capabilities of the master and sub-agents. Finally, the master agent feeds the optimization results back to the console via the HAI protocol.

Together, HAI, UAP, and MEK form a complete collaborative loop within the Co-TAP agent protocol architecture. HAI provides real-time process visibility and a channel for human intervention. UAP provides the foundation for connection and communication, supporting task delegation and data flow between heterogeneous agents. MEK then consolidates the high-value experience generated during collaboration into system knowledge, which is distributed and absorbed through the agent network built by UAP, ultimately achieving the continuous improvement of collective intelligence based on experiential learning.
\section{Conclusion and Outlook}
\subsection{Conclusion}
This report has detailed the Co-TAP agent protocol architecture, proposed for large model agent technology projects. Through the meticulous design of three core protocols—HAI, UAP, and MEK—this framework successfully constructs a full-stack agent solution covering infrastructure connectivity, real-time human-agent interaction, and collective cognitive learning. The core value of this protocol architecture lies in fundamentally enhancing the interoperability of multi-agent systems, the trustworthiness of interactions, and the continuity of cognitive learning. Specifically, the HAI protocol, acting as a bridge between the agent and the user interface, enables real-time, trustworthy, and synergistic human-agent interaction. Its core lies in a unified event stream architecture and Server-Sent Events (SSE), which support token-by-token streaming, providing an extremely responsive user experience. HAI ensures that users can control the agent's entire lifecycle and seamlessly embeds Human-in-the-Loop (HITL) processes through Tool Call Events, guaranteeing the correctness of critical steps and the trustworthiness of the interaction. The UAP protocol addresses two core problems in multi-agent systems: the inability of agents to discover each other autonomously, and the high costs of cross-protocol adaptation due to the coexistence of multiple protocols. It adopts a core strategy of modular decomposition and ecosystem-oriented construction, providing a unified service plane. Through the protocol conversion engine built into the UAP gateway (using the Adapter Pattern), it achieves lossless conversion and interoperability between heterogeneous protocols (such as A2A, ACP, MCP, etc.), reducing the development costs of heterogeneous communication. Furthermore, UAP effectively reduces the extra overhead caused by redundant context transmission in agent chains through on-demand disclosure and sensitive data masking strategies. Finally, the MEK protocol establishes the core M-E-K (Memory-Extraction-Knowledge) logic chain, resolving the bottleneck of difficulties in sharing knowledge and experience. By defining structured memories (such as Workflow Memory) and supporting multi-layered storage, MEK provides a standardized memory management framework for agents. Its Extraction (E) stage follows a strict, standardized four-step process to distill personalized memories into universal KnowledgeItems. Through conflict detection and fusion mechanisms in the knowledge absorption process, MEK enables the cross-agent reuse and internalization of knowledge within the agent ecosystem, driving the continuous improvement of collective intelligence based on experiential learning.
\subsection{Outlook}
Looking ahead, the primary task for the Co-TAP agent protocol architecture is to strengthen its protocol extensibility to adapt to the continuous emergence of new communication protocols and domain-specific protocols. The HAI protocol will be extended on the basis of the existing event-driven architecture to support more complex interaction paradigms and higher-level user intent feedback mechanisms. The UAP registry will also continuously optimize its protocol extension mechanism, supporting the rapid integration of new protocols by adding protocol types and protocol-specific extension fields. To reduce the difficulty of integrating support for new protocols, the plug-in architecture of the UAP gateway will be enhanced, and toolkits will be developed to automate the generation of adapter modules. On the cognitive level, the memory structure of the MEK protocol will be generalized and extended to more easily support new types of multimodal data and emerging memory types, thereby addressing more complex cognitive demands.

\section{Authors}

\textbf{HAI:} Shunyu An, Miao Wang, Yongchao Li, Dong Wan, Lina Wang, Ling Qin, Liqin Gao, Congyao Fan

\textbf{UAP:} Zhiyong Mao, Jiange Pu, Wenji Xia, Dong Zhao, Zhaohui Hao, Rui Hu

\textbf{MEK:} Ji Lu, Guiyue Zhou, Baoyu Tang

\textbf{Overall conception and coordination:} Yanqin Gao, Yongsheng Du, Daigang Xu, Lingjun Huang, Baoli Wang, Xiwen Zhang, Luyao Wang, Shilong Liu
\bibliographystyle{unsrt}  
\bibliography{references}

\begin{thebibliography}{10}

\bibitem{xi2025rise}
Zhiheng Xi, Wenxiang Chen, Xin Guo, Wei He, Yiwen Ding, Boyang Hong, Ming Zhang, Junzhe Wang, Senjie Jin, Enyu Zhou, et~al.
\newblock The rise and potential of large language model based agents: A survey.
\newblock {\em Science China Information Sciences}, 68(2):121101, 2025.

\bibitem{wang2024survey}
Lei Wang, Chen Ma, Xueyang Feng, Zeyu Zhang, Hao Yang, Jingsen Zhang, Zhiyuan Chen, Jiakai Tang, Xu~Chen, Yankai Lin, et~al.
\newblock A survey on large language model based autonomous agents.
\newblock {\em Frontiers of Computer Science}, 18(6):186345, 2024.

\bibitem{hao2023reasoning}
Shibo Hao, Yi~Gu, Haodi Ma, Joshua~Jiahua Hong, Zhen Wang, Daisy~Zhe Wang, and Zhiting Hu.
\newblock Reasoning with language model is planning with world model.
\newblock {\em arXiv preprint arXiv:2305.14992}, 2023.

\bibitem{krishnan2025advancing}
Naveen Krishnan.
\newblock Advancing multi-agent systems through model context protocol: Architecture, implementation, and applications.
\newblock {\em arXiv preprint arXiv:2504.21030}, 2025.

\bibitem{mushtaq2025harnessing}
Abdullah Mushtaq, Rafay Naeem, Ibrahim Ghaznavi, Imran Taj, Imran Hashmi, and Junaid Qadir.
\newblock Harnessing multi-agent llms for complex engineering problem-solving: A framework for senior design projects.
\newblock In {\em 2025 IEEE Global Engineering Education Conference (EDUCON)}, pages 1--10. IEEE, 2025.

\bibitem{wasif2025multi}
Mubeen Wasif and David Tunkel.
\newblock Multi-agent collaboration in ai: Enhancing software development with autonomous llms.
\newblock {\em Preprint at https://doi. org/10.13140/RG}, 2(31588.08328), 2025.

\bibitem{deloach2005multiagent}
Scott~A DeLoach.
\newblock Multiagent systems engineering of organization-based multiagent systems.
\newblock {\em ACM SIGSOFT Software Engineering Notes}, 30(4):1--7, 2005.

\bibitem{zhang2025collective}
Terry~Jingchen Zhang, Yongjin Yang, Yinya Huang, Sirui Lu, Bernhard Sch{\"o}lkopf, and Zhijing Jin.
\newblock Collective intelligence: On the promise and reality of multi-agent systems for ai-driven scientific discovery.
\newblock {\em Preprints, August}, 2025.

\bibitem{ghareeb2025robin}
Ali~Essam Ghareeb, Benjamin Chang, Ludovico Mitchener, Angela Yiu, Caralyn~J Szostkiewicz, Jon~M Laurent, Muhammed~T Razzak, Andrew~D White, Michaela~M Hinks, and Samuel~G Rodriques.
\newblock Robin: A multi-agent system for automating scientific discovery.
\newblock {\em arXiv preprint arXiv:2505.13400}, 2025.

\bibitem{tebourbi2025bpmn}
Hedi Tebourbi, Sana Nouzri, Yazan Mualla, Meryem El~Fatimi, Amro Najjar, Abdeljalil Abbas-Turki, and Mahjoub Dridi.
\newblock Bpmn-based design of multi-agent systems: Personalized language learning workflow automation with rag-enhanced knowledge access.
\newblock {\em Information}, 16(9):809, 2025.

\bibitem{sulis2022multi}
Emilio Sulis and Kuldar Taveter.
\newblock Multi-agent systems and business process management.
\newblock In {\em Agent-Based Business Process Simulation: A Primer with Applications and Examples}, pages 131--140. Springer, 2022.

\bibitem{li2008agent}
Xiaochen Li, Wenji Mao, Daniel Zeng, and Fei-Yue Wang.
\newblock Agent-based social simulation and modeling in social computing.
\newblock In {\em International Conference on Intelligence and Security Informatics}, pages 401--412. Springer, 2008.

\bibitem{davidsson2000multi}
Paul Davidsson.
\newblock Multi agent based simulation: beyond social simulation.
\newblock In {\em International workshop on multi-agent systems and agent-based simulation}, pages 97--107. Springer, 2000.

\bibitem{mascardi2019engineering}
Viviana Mascardi, Danny Weyns, Alessandro Ricci, Clara~Benac Earle, Arthur Casals, Moharram Challenger, Amit Chopra, Andrei Ciortea, Louise~A Dennis, {\'A}lvaro~Fern{\'a}ndez D{\'\i}az, et~al.
\newblock Engineering multi-agent systems: State of affairs and the road ahead.
\newblock {\em ACM SIGSOFT Software Engineering Notes}, 44(1):18--28, 2019.

\bibitem{yan2025beyond}
Bingyu Yan, Zhibo Zhou, Litian Zhang, Lian Zhang, Ziyi Zhou, Dezhuang Miao, Zhoujun Li, Chaozhuo Li, and Xiaoming Zhang.
\newblock Beyond self-talk: A communication-centric survey of llm-based multi-agent systems.
\newblock {\em arXiv preprint arXiv:2502.14321}, 2025.

\bibitem{lazaridou2020emergent}
Angeliki Lazaridou and Marco Baroni.
\newblock Emergent multi-agent communication in the deep learning era.
\newblock {\em arXiv preprint arXiv:2006.02419}, 2020.

\bibitem{ouyang2025code2mcp}
Chaoqian Ouyang, Ling Yue, Shimin Di, Libin Zheng, Shaowu Pan, and Min-Ling Zhang.
\newblock Code2mcp: A multi-agent framework for automated transformation of code repositories into model context protocol services.
\newblock {\em arXiv preprint arXiv:2509.05941}, 2025.

\bibitem{qasim2024effective}
Awais Qasim, Arslan Ghouri, and Adeel Munawar.
\newblock An effective approach for reducing data redundancy in multi-agent system communication.
\newblock {\em Multiagent and Grid Systems}, 20(1):69--88, 2024.

\bibitem{cemri2025multi}
Mert Cemri, Melissa~Z Pan, Shuyi Yang, Lakshya~A Agrawal, Bhavya Chopra, Rishabh Tiwari, Kurt Keutzer, Aditya Parameswaran, Dan Klein, Kannan Ramchandran, et~al.
\newblock Why do multi-agent llm systems fail?
\newblock {\em arXiv preprint arXiv:2503.13657}, 2025.

\bibitem{julian2004developing}
Vicente Julian and Vicent Botti.
\newblock Developing real-time multi-agent systems.
\newblock {\em Integrated Computer-Aided Engineering}, 11(2):135--149, 2004.

\bibitem{menda2018deep}
Kunal Menda, Yi-Chun Chen, Justin Grana, James~W Bono, Brendan~D Tracey, Mykel~J Kochenderfer, and David Wolpert.
\newblock Deep reinforcement learning for event-driven multi-agent decision processes.
\newblock {\em IEEE Transactions on Intelligent Transportation Systems}, 20(4):1259--1268, 2018.

\bibitem{meyer2014event}
Ruth Meyer.
\newblock Event-driven multi-agent simulation.
\newblock In {\em International Workshop on Multi-Agent Systems and Agent-Based Simulation}, pages 3--16. Springer, 2014.

\bibitem{calvaresi2017challenge}
Davide Calvaresi, Mauro Marinoni, Arnon Sturm, Michael Schumacher, and Giorgio Buttazzo.
\newblock The challenge of real-time multi-agent systems for enabling iot and cps.
\newblock In {\em Proceedings of the international conference on web intelligence}, pages 356--364, 2017.

\bibitem{venkadesh2024unlocking}
P~Venkadesh, SV~Divya, and K~Subash Kumar.
\newblock Unlocking ai creativity: A multi-agent approach with crewai.
\newblock {\em Journal of Trends in Computer Science Smart Technology}, 6(4):338--356, 2024.

\bibitem{jeong2025study}
Cheonsu Jeong.
\newblock A study on the mcp x a2a framework for enhancing interoperability of llm-based autonomous agents.
\newblock {\em arXiv preprint arXiv:2506.01804}, 2025.

\bibitem{cao2025large}
Pengfei Cao, Tianyi Men, Wencan Liu, Jingwen Zhang, Xuzhao Li, Xixun Lin, Dianbo Sui, Yanan Cao, Kang Liu, and Jun Zhao.
\newblock Large language models for planning: A comprehensive and systematic survey.
\newblock {\em arXiv preprint arXiv:2505.19683}, 2025.

\bibitem{gorodetsky2020system}
Vladimir Gorodetsky, Petr Skobelev, and Vladim{\i}r Marik.
\newblock System engineering view on multi-agent technology for industrial applications: Barriers and prospects.
\newblock {\em Cybernetics and Physics}, 9(1):13--30, 2020.

\bibitem{ehtesham2025survey}
Abul Ehtesham, Aditi Singh, Gaurav~Kumar Gupta, and Saket Kumar.
\newblock A survey of agent interoperability protocols: Model context protocol (mcp), agent communication protocol (acp), agent-to-agent protocol (a2a), and agent network protocol (anp).
\newblock {\em arXiv preprint arXiv:2505.02279}, 2025.

\bibitem{hosseini2025role}
Soodeh Hosseini and Hossein Seilani.
\newblock The role of agentic ai in shaping a smart future: A systematic review.
\newblock {\em Array}, page 100399, 2025.

\bibitem{baldoni2020fragility}
Matteo Baldoni, Cristina Baroglio, and Roberto Micalizio.
\newblock Fragility and robustness in multiagent systems.
\newblock In {\em International Workshop on Engineering Multi-Agent Systems}, pages 61--77. Springer, 2020.

\bibitem{jang2004efficient}
Myeong-Wuk Jang, Amr Ahmed, and Gul Agha.
\newblock Efficient agent communication in multi-agent systems.
\newblock In {\em International Workshop on Software Engineering for Large-Scale Multi-agent Systems}, pages 236--253. Springer, 2004.

\bibitem{wang2023cooperative}
Xin Wang, Chen Zhao, Tingwen Huang, Prasun Chakrabarti, and J{\"u}rgen Kurths.
\newblock Cooperative learning of multi-agent systems via reinforcement learning.
\newblock {\em IEEE Transactions on Signal and Information Processing over Networks}, 9:13--23, 2023.

\bibitem{dorri2018multi}
Ali Dorri, Salil~S Kanhere, and Raja Jurdak.
\newblock Multi-agent systems: A survey.
\newblock {\em Ieee Access}, 6:28573--28593, 2018.

\bibitem{ghafarollahi2025sciagents}
Alireza Ghafarollahi and Markus~J Buehler.
\newblock Sciagents: automating scientific discovery through bioinspired multi-agent intelligent graph reasoning.
\newblock {\em Advanced Materials}, 37(22):2413523, 2025.

\bibitem{banares2005multi}
R~Banares-Alcantara, L~Jim{\'e}nez, and A~Aldea.
\newblock Multi-agent systems for ontology-based information retrieval.
\newblock In {\em Computer Aided Chemical Engineering}, volume~20, pages 1549--1554. Elsevier, 2005.

\bibitem{yu2012group}
Junyan Yu and Long Wang.
\newblock Group consensus of multi-agent systems with directed information exchange.
\newblock {\em International Journal of Systems Science}, 43(2):334--348, 2012.

\bibitem{su2001logical}
Kaile Su, Xudong Luo, Huaiqing Wang, Chengqi Zhang, Shichao Zhang, and Qingfeng Chen.
\newblock A logical framework for knowledge sharing in multi-agent systems.
\newblock In {\em International Computing and Combinatorics Conference}, pages 561--570. Springer, 2001.

\bibitem{liu2022distributed}
Shicheng Liu and Minghui Zhu.
\newblock Distributed inverse constrained reinforcement learning for multi-agent systems.
\newblock {\em Advances in Neural Information Processing Systems}, 35:33444--33456, 2022.

\bibitem{baldoni2023accountability}
Matteo Baldoni, Cristina Baroglio, Roberto Micalizio, and Stefano Tedeschi.
\newblock Accountability in multi-agent organizations: from conceptual design to agent programming.
\newblock {\em Autonomous Agents and Multi-Agent Systems}, 37(1):7, 2023.

\bibitem{yousefli2020maintenance}
Zahra Yousefli, Fuzhan Nasiri, and Osama Moselhi.
\newblock Maintenance workflow management in hospitals: An automated multi-agent facility management system.
\newblock {\em Journal of Building Engineering}, 32:101431, 2020.

\bibitem{sarkar2025survey}
Anjana Sarkar and Soumyendu Sarkar.
\newblock Survey of llm agent communication with mcp: A software design pattern centric review.
\newblock {\em arXiv preprint arXiv:2506.05364}, 2025.

\bibitem{newton2021scalability}
Charles Newton, John Singleton, Cameron Copland, Sarah Kitchen, and Jeffrey Hudack.
\newblock Scalability in modeling and simulation systems for multi-agent, ai, and machine learning applications.
\newblock In {\em Artificial Intelligence and Machine Learning for Multi-Domain Operations Applications III}, volume 11746, pages 534--552. SPIE, 2021.

\bibitem{partalas2008hybrid}
Ioannis Partalas, Ioannis Feneris, and Ioannis Vlahavas.
\newblock A hybrid multiagent reinforcement learning approach using strategies and fusion.
\newblock {\em International Journal on Artificial Intelligence Tools}, 17(05):945--962, 2008.

\bibitem{ding2024learning}
Shifei Ding, Wei Du, Ling Ding, Lili Guo, and Jian Zhang.
\newblock Learning efficient and robust multi-agent communication via graph information bottleneck.
\newblock In {\em Proceedings of the AAAI Conference on Artificial Intelligence}, volume~38, pages 17346--17353, 2024.

\bibitem{lee1998stability}
Lyndon~C Lee, Hyacinth~S Nwana, Divine~T Ndumu, and Philippe De~Wilde.
\newblock The stability, scalability and performance of multi-agent systems.
\newblock {\em BT Technology Journal}, 16(3):94--103, 1998.

\bibitem{el2009multi}
ABEER El-Korany and KHALED EL-Bahnasy.
\newblock A multi-agent framework to facilitate knowledge sharing.
\newblock {\em Journal of Artificial Intelligence}, 2(1):17--28, 2009.

\end{thebibliography}

\end{document}